\documentclass{article}

\usepackage{arxiv}
\usepackage[numbers]{natbib}
\usepackage[utf8]{inputenc} 
\usepackage[T1]{fontenc}  
\usepackage{hyperref}       
\usepackage{url}            
\usepackage{booktabs}       
\usepackage{amsfonts}       
\usepackage{nicefrac}       
\usepackage{microtype}      
 
\usepackage{graphicx}
\graphicspath{ {./images/} }

\title{An Item Response Theory-based R module for Algorithm Portfolio Analysis}

\author{
 Brodie Oldfield \\
  CSIRO's Data61\\
  Eveleigh, NSW, 2015, Australia \\
  \texttt{brodie.oldfield@data61.csiro.au} \\
   \And
 Sevvandi Kandanaarachchi \\
  CSIRO's Data61\\
  Clayton VIC, 3068, Australia \\
  \And
 Ziqi Xu \\
 CSIRO's Data61\\
  Clayton VIC, 3068, Australia \\
  \And
 Mario Andrés Muñoz \\
   School of Computer and Information Systems\\
  The University of Melbourne\\
  Parkville, VIC, 3010, Australia \&\\
  ARC Centre in Optimisation Technologies, Integrated Methodologies, and Applications (OPTIMA)\\
  Carlton, VIC, 3052, Australia \\
}

\begin{document}
\maketitle
\begin{abstract}
Experimental evaluation is crucial in AI research, especially for assessing algorithms across diverse tasks. Many studies often evaluate a limited set of algorithms, failing to fully understand their strengths and weaknesses within a comprehensive portfolio. This paper introduces an Item Response Theory (IRT) based analysis tool for algorithm portfolio evaluation called AIRT-Module. Traditionally used in educational psychometrics, IRT models test question difficulty and student ability using responses to test questions. Adapting IRT to algorithm evaluation, the AIRT-Module contains a Shiny web application and the R package \texttt{airt}. AIRT-Module uses algorithm performance measures to compute anomalousness, consistency, and difficulty limits for an algorithm and the difficulty of test instances. The strengths and weaknesses of algorithms are visualised using the difficulty spectrum of the test instances. AIRT-Module offers a detailed understanding of algorithm capabilities across varied test instances, thus enhancing comprehensive AI method assessment. It is available at \url{https://sevvandi.shinyapps.io/AIRT/}. 
\end{abstract}

\keywords{Item Response Theory \and Algorithm Evaluation \and Machine Learning \and R Language \and R Shiny \and Benchmarking Portfolios}

\section{Motivation and significance}
\paragraph{}
Experimental evaluation is critical for AI research, especially for problems with elusive theoretical evaluation. AI researchers are interested in the performance of a particular method for a specific problem instance, across multiple instances, and against other methods. Evaluating a diverse set of algorithms across a comprehensive set of test instances contributes to an increased understanding of the interplay between instance characteristics, algorithm mechanisms, and algorithm performance. Such an evaluation helps determine an algorithm's strengths and weaknesses and provides a broad overview of the collective capabilities of an algorithm portfolio. However, many studies that evaluate only a small number of algorithms on a limited set of test instances fail to reveal where any algorithm belongs within a state-of-the-art algorithm portfolio's capabilities or where algorithms' unique strengths and weaknesses lie when considering a diverse range of test problem difficulties and challenges. In this paper, we present AIRT-Module, an Item Response Theory (IRT)-based analysis tool for evaluating a portfolio of algorithms.

\paragraph{}
IRT \cite{embretson2013item,RonaldKHambleton1985} is commonly used in educational psychometrics to analyse and model responses to test questions. The premise of IRT is that there is an underlying characteristic, such as verbal/mathematical ability, boredom proneness \cite{Struk2017} or misinformation susceptibility \cite{Maertens2024} that is difficult to measure directly but can be modelled via responses to questions. In educational psychometrics, 2-parameter IRT models use marks for a group of test questions to evaluate a question's difficulty and discrimination and a student's ability. Different IRT models are appropriate depending on the type of response: dichotomous, polytomous and continuous. Dichotomous models are used for binary responses such as true/false questions. Polytomous models are used for discrete-valued, ordinal responses commonly seen in surveys ranging from ``strongly agree'' to ``strongly disagree'' options. Continuous models are used for continuous-valued responses such as extended written responses in exams~\citep{embretson2013item}. While research into educational psychometrics has used IRT since the 1960s, its use in machine learning is more recent~\cite{MARTINEZPLUMED201918}. The Algorithmic Item Response Theory (AIRT) framework is one such adaptation~\cite{JMLR2023SKKSM}. AIRT-Module comprises an R package called \texttt{airt} and a Shiny web application called the AIRT Shiny App. Using algorithm performance values as input, AIRT-Module computes an algorithm's anomalousness, consistency and difficulty limit, and the test instance difficulty. For a given problem set, the space of test instance difficulties constitutes the problem difficulty spectrum.

\begin{itemize}
    \item \textbf{Anomalousness} is a boolean value flagged if an algorithm excels with difficult problems but struggles with easy problems.
    \item \textbf{Consistency} is a numeric value that indicates the stability of the performance. A low consistency algorithm gives fluctuating performance for datasets of similar difficulty, whereas a high consistency algorithm gives similar performance irrespective of dataset difficulty.
    \item \textbf{Difficulty Limit} is a numeric value that describes the highest difficulty level an algorithm can handle. A higher difficulty limit score means that the algorithm can handle harder problems.
    \item \textbf{Problem difficulty spectrum} is the one-dimensional space where test instance difficulty values reside, ranging from easy to hard. 
\end{itemize}

In the \texttt{airt} R package, these attributes can be computed and plotted using continuous data, e.g., an algorithm's accuracy score, or polytomous (discrete) data, e.g. a grading system between A and F. The Shiny App introduces a dynamic user interface to \texttt{airt} with the functionality to upload datasets, transform datasets, change function parameters, and download the resultant plots. We developed the AIRT Shiny App to increase its availability to users unfamiliar with the R language.

\section{Software description}
\begin{figure}[!ht]
    \centering
    \includegraphics[width=\textwidth]{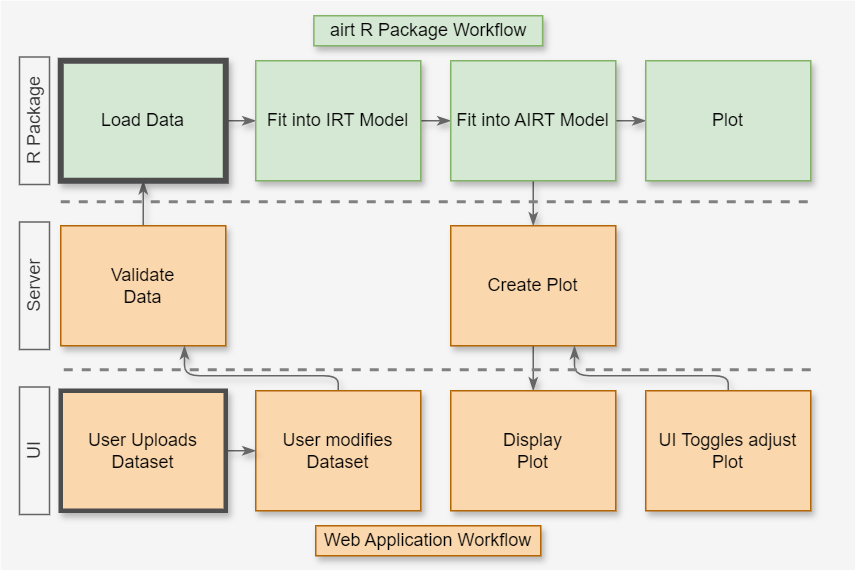}
    \caption{AIRT-Module workflows: One (Green) showing the workflow when only using the \texttt{airt} R Package, the other (Orange) when using the AIRT Shiny App.  Thick-bordered cells indicate starting actions in the workflow.}
    \label{fig:pipeline}
\end{figure}

The AIRT-Module operates under an Input $\Rightarrow$ Model $\Rightarrow$ Output system. The input is a dataset of performance values for a portfolio of algorithms to a diverse set of test instances. An IRT model adapted for algorithm evaluation is fitted to this data~\cite{JMLR2023SKKSM}. We will call this the AIRT model for the remainder of the paper. The output is the resulting model, its parameters and the created plots. While the R package has the functionality to fit polytomous performance data, our focus here is on continuous data such as classification accuracies (ranging in $[0, 100]$).

\begin{figure}[!ht]
    \centering
    \includegraphics[width=\textwidth]{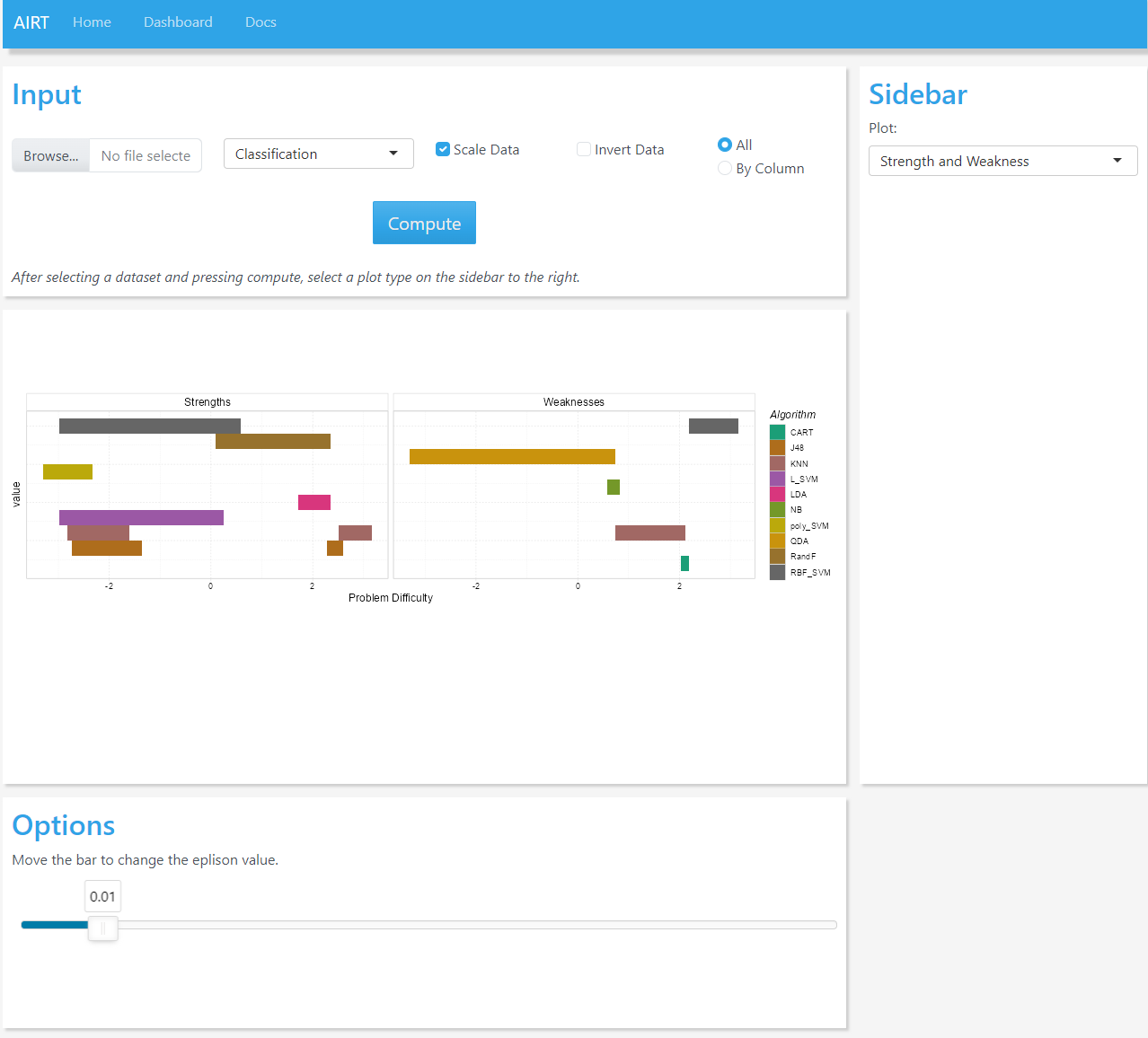}
    \caption{Dashboard view of the AIRT Shiny App.}
    \label{fig:Dashboard}
\end{figure}

The AIRT Shiny App is built using the \texttt{airt} R package and has two interfaces: a walkthrough interface and a dashboard (See Figure~\ref{fig:Dashboard}). The walkthrough interface is oriented in a presentation manner, where users are shown \texttt{airt} visualisations and analysis as sections. Each section is only rendered when a user chooses to continue and contains UI elements allowing plots to respond dynamically to user inputs. Each section includes an explanation of the plot and critical methods of analysis. In contrast, the dashboard interface generates all plots simultaneously and renders a plot at a time based on user preference. 

\subsection{Software architecture}
\subsubsection{Overview}
Figure~\ref{fig:pipeline} illustrates, at a high level, the AIRT-Module architecture, consisting of the R Package \texttt{airt}, which handles IRT and related computations, and the AIRT Shiny App made using R Shiny, which runs the R Package and renders the results to users. Users can run the \texttt{airt} R Package independently of the AIRT Shiny App.

\subsubsection{R Package}
After suitably mapping algorithms and test problems to the IRT setting~\cite{JMLR2023SKKSM}, the R packages \texttt{mirt}~\cite{mirtR} and  \texttt{EstCRM}~\cite{estcrmR} are used to fit an AIRT Model for polytomous and continuous data respectively. To allow for a broader range of algorithms, such as anomalous algorithms,  parts of \texttt{EstCRM} code were modified~\cite{JMLR2023SKKSM}. After fitting the model, AIRT-Module computes algorithm attributes and finds the strengths and weaknesses using \texttt{latent\_trait\_analysis()}. The \texttt{airt} attributes are then used within \texttt{autoplot} to create plots.

\subsubsection{AIRT Shiny App}
AIRT Shiny App uses Shiny by posit, allowing a server to run R code and communicate with a user's session. Structurally, the AIRT\_Shiny project directory contains a \texttt{UI.R} document, which houses the HTML/CSS/JavaScript the web page is scaffolded from, a \texttt{server.R} document, which contains the main rendering and logic functions, and utility documents to group related functions. Pre-computed datasets are under the ./Data directory and are loaded into \texttt{server.R} on server startup.

AIRT Shiny App also has an Input, Model, and Output pipeline. Inputs are datasets where users can upload their dataset in CSV format or use a pre-generated example dataset. The dataset is validated and modelled within relevant \texttt{airt} functions such as \texttt{cirtmodel()} and \texttt{latent\_trait\_analysis()}. Where possible and appropriate, the output of functions is cached. Plots are generated from the outputs and rendered by the UI.

\subsubsection{Deployment}
A release version of \texttt{airt} is available from the Comprehensive R Archive Network (CRAN) repository, while a development version is available in GitHub. Users can access the AIRT Shiny App at \url{https://sevvandi.shinyapps.io/AIRT/}, deploy it locally via RStudio or host it using a service that handles Shiny-compatible environments.
 
\begin{verbatim}
//CRAN Release Version
install.packages("airt")

//GitHub Development Version
install.packages("devtools")
devtools::install_github("sevvandi/airt")
\end{verbatim}
Installation instructions for AIRT Shiny App are available on \url{https://github.com/broldfield/AIRT\_Shiny} under the \texttt{README.md} file.

\subsection{Software Functionality}

\subsubsection{AIRT R Package Functions}
Data is loaded into the IRT model using \texttt{cirtmodel()} for continuous data or \texttt{pirtmodel()} for polytomous data. Typically, functions suffixed or prefixed with `c' or `crm' are for continuous data, and functions with `p' are for polytomous data. These functions accept a dataset as a data frame and output the IRT parameters relevant to \texttt{airt}. For \texttt{cirtmodel()}, this would be in the element \texttt{cirtmodel\_output\$model\$param}.

The original data frame, the param element, and an epsilon value are used as parameters in \texttt{latent\_trait\_analysis()} to create the \texttt{airt} attributes. This output, denoted as LTA\_output, is used in all \texttt{airt} plotting functions for analysis.

The \texttt{autoplot(object, plottype)} function is used to generate a plot using \texttt{ggplot} \cite{ggplot2} based on the input object and the plot type specified. For LTA\_output, there are four plot types of relevance. Plot types 1 and 2 show Performance against Problem Difficulty as scatter plots. Plot type 3 shows smoothing splines fitted to the performance values where the $x$ axis denotes the problem difficulty. The smoothing splines are particularly important as the best-performing algorithm for a given problem difficulty will be the one whose spline is at the top. Type 4 generates a bar chart version of plot type 3, where the default setting corresponding to \texttt{epsilon} = 0 shows the best algorithm for every value in the problem difficulty spectrum. 
The \texttt{epsilon} value is a goodness threshold. When \texttt{epsilon}  = 0, only the best algorithm for every problem difficulty value is considered. When \texttt{epsilon} = 0.01, algorithms with performance within 0.01 of the best are considered. 
Modifying the \texttt{epsilon} value in \texttt{latent\_trait\_analysis()} allows multiple algorithms to overlap in the same problem difficulty.

Heatmaps can be generated for continuous data using the \texttt{heatmaps\_crm}, showing positive sloped lines if an algorithm is not anomalous, thinner lines for more discriminating algorithms and blurrier lines for more consistent algorithms.

A user can analyse whether the fitted AIRT model is appropriate by employing \texttt{model\_goodness\_crm()} and \texttt{effectiveness\_crm()}. When the output of \texttt{model\_goodness\_crm()} is passed to \texttt{autoplot}, the distribution of errors is plotted. The output of \texttt{effectiveness\_crm()} and a plot type are used within \texttt{autoplot} to create three different plots. Type 1: Actual Effectiveness against Effectiveness Tolerance, Type 2: Predicted Effectiveness against Effective Tolerance, and Type 3: Predicted Effectiveness against Actual Effectiveness. Type 3 is important as the closer the points are to the dotted line $y = x$, the better the fitted AIRT model.

For polytomous data, the output of \texttt{pirtmodel()} is used with \texttt{tracelines\_poly()} and \texttt{autoplot} to create tracelines showing the probability of reaching a performance band. Performance bands are labelled 1 to 5, with the probability of scoring 5 being higher for easier datasets and lower for challenging datasets.
Similar to how model goodness is visualised for continuous data, \texttt{model\_goodness\_poly()} and \texttt{effectiveness\_poly()} display the same plots and use the same plot types. 

\subsubsection{AIRT Shiny App Functions}
Users can use a pre-generated example file or upload their dataset as a CSV document. When a user uploads a dataset to the server, a validation check is committed over the whole CSV to ensure \texttt{cirtmodel()} can use it. The validation primarily checks that all fields besides the column names are numeric. 

As the AIRT Shiny App aims to assist in data analysis, additional tools exist to modify the dataset. Modifying the dataset occurs before the dataset is processed, with UI elements allowing the user to: 

\begin{itemize}
    \item `Scale Data' which fits each dataset value to be a proportion between 0 and 1 by flagging \texttt{scale = TRUE} in \texttt{latent\_trait\_analysis()}. 
    
    \item `Invert Data' transforms the dataset using $\max{x} - x$ for each column to map low to high values. This functionality is needed when low values indicate better performance, such as when root mean square error is the performance metric.  
    
    \item `Scale By' determines whether the proportion of a value received from `Scale Data' is calculated per Column (Algorithm) or over the whole dataset.
\end{itemize}

By default, a CSV's minimum and maximum performance values are validated to be between 0 and 1. If performance values are not scaled, e.g. watts, then users can untick `Scale Data'. This property is set within \texttt{cirtmodel()} and \texttt{latent\_trait\_analysis()} as an optional parameters \texttt{min.item} and \texttt{max.item}. 

\begin{figure}[!ht]
    \centering
    \includegraphics[width=100mm]{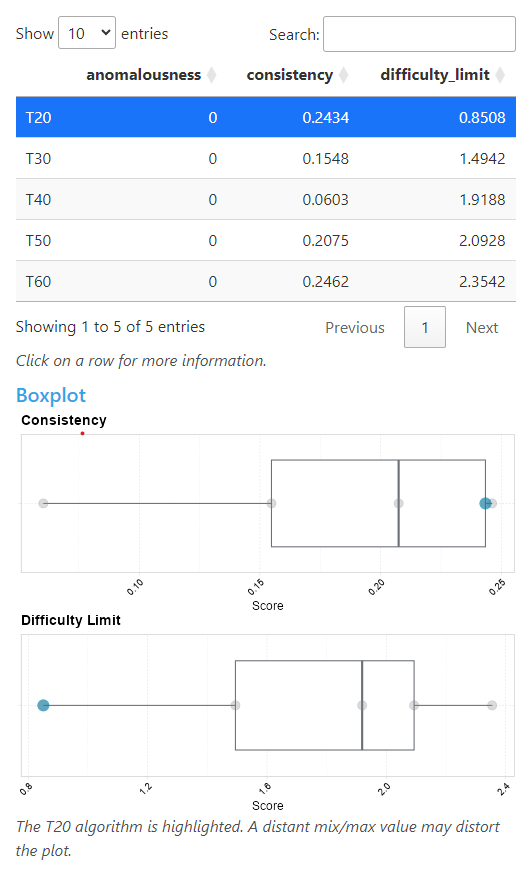}
    \caption{A table showing the \texttt{airt} attributes of an algorithm portfolio. Boxplots show its consistency and difficulty when an algorithm is highlighted on the table.}
    \label{Boxplot image}
\end{figure}

Furthering the data analysis tools found in \texttt{airt} are plots and tables unique to the AIRT Shiny App, which expand upon existing data presented to the user:

\begin{itemize}
    \item When a user selects an algorithm when viewing the AIRT Attributes table, the Difficulty Limit and Consistency data in \texttt{latent\_trait\_analysis} is used to create a box plot (Figure \ref{Boxplot image}). This box plot shows all the algorithms as points with the selected algorithm highlighted.
    \item Extending from the Strengths and Weaknesses bar chart, we can compute the proportion of the latent trait spectrum occupied by each algorithm. 
    The table containing these proportions updates alongside the epsilon slider next to the bar chart (Figure \ref{Splines&SW}).
\end{itemize}

Users can download their generated plots and tables in PNG format inside a tar file, generated by \texttt{downloadHandler()} and create temp directories for that session.

\begin{figure}[!ht]
    \centering
    \includegraphics[width=100mm]{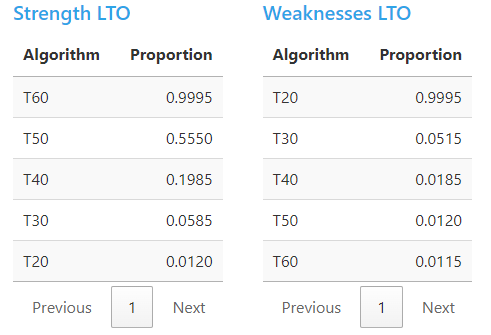}
    \caption{A table generated under the Strengths and Weaknesses Bar Chart showing the proportion occupied by an algorithm on the problem difficulty spectrum for the selected epsilon value.}
    \label{LTO image}
\end{figure}

\section{Illustrative examples}

We follow the workflows shown in Figure \ref{fig:pipeline} from the AIRT Shiny App and R Package perspectives to complete the task of determining the strengths and weaknesses of the algorithm portfolio.

\subsection{R Package Workflow}
Firstly, we would load the \texttt{airt} library and load in our data.

\begin{verbatim}
library("airt")
data("classification_cts")
df <- classification_cts
\end{verbatim}

Our data, pre-supplied by the \texttt{airt} package, can be replaced by a user's data conforming to the expected format. The pre-supplied data is taken from the MATILDA data repository~\cite{matilda}.

\begin{verbatim}
irt_params <- cirtmodel(df)
airt_params <- latent_trait_analysis(df, 
    paras = irt_params$model$param,
    epsilon = 0)
\end{verbatim}

The data frame is passed to \texttt{cirtmodel()} to fit the AIRT model. The IRT parameters stored in \texttt{param} are then passed into \texttt{latent\_trait\_analysis()} alongside the original data frame with an epsilon value. The plots are generated with a default \texttt{epsilon} value of 0, which shows the strongest and weakest algorithms for every value in the problem difficulty spectrum. Suppose \texttt{epsilon} is incremented by 0.1. In that case, algorithms within the strongest and weakest by 0.1 in performance are also displayed, with the range of displayed algorithms increasing with the epsilon value.

\begin{verbatim}
autoplot(airt_params, plottype = 3)
autoplot(airt_params, plottype = 4)
\end{verbatim}

Four plots can be generated from \texttt{airt\_params} using \texttt{autoplot} by setting the value of \texttt{plottype}. Options \texttt{\{1,2\}} plot algorithm performance with problem difficulty spectrum on the $x$ axis and algorithm performance on the $y$ axis. Option \texttt{3} displays smoothing splines fitted to the performance values as a function of problem difficulty. The splines corresponding to the strongest algorithm for a given problem difficulty come at the top, while the weakest come at the bottom. Option \texttt{4} displays the strengths and weaknesses of algorithms across the problem spectrum.

\subsection{AIRT Shiny App Workflow}
\begin{figure}[!ht]
    \centering
    \includegraphics[width=125mm]{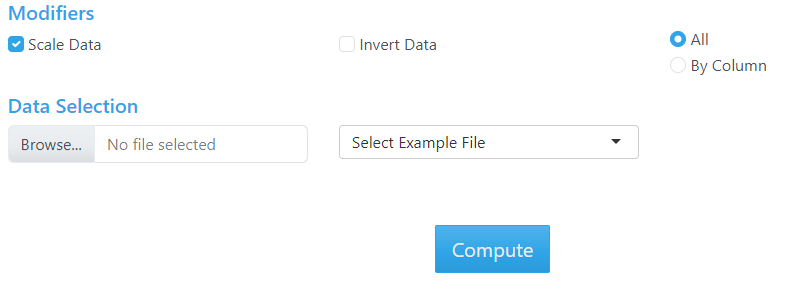}
    \caption{Inputs shown at the start of AIRT Shiny App. Modifiers transform the dataset, Data Selection allows users to either upload a dataset or select an example dataset from MATILDA.}
    \label{inputs image}
\end{figure}

Here, we need a dataset to analyse, similar to the \texttt{airt} package. As shown in Figure \ref{inputs image}, AIRT Shiny App has controls for uploading or selecting datasets. In this workflow example, we will use the `Classification' example dataset from the `Select Example File' Dropdown box. Alternatively, the user could use the `Browse' file upload input to upload their dataset as a CSV to the application. If the user chooses to upload their dataset, it goes through an additional validation check.

\begin{figure}[!ht]
    \centering
    \includegraphics[width=125mm]{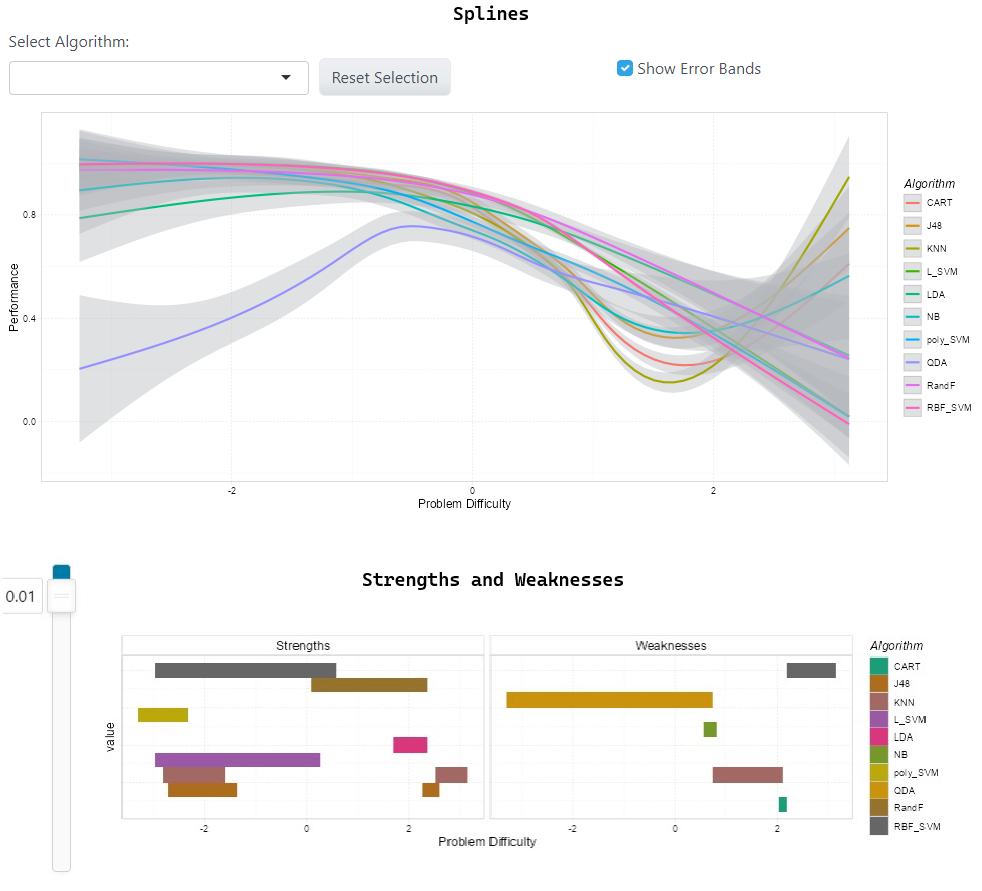}
    \caption{Splines Plot (Above), Strengths and Weaknesses Plot (Below). Only the AIRT Shiny App has the UI controls.}
    \label{Splines&SW}
\end{figure}

After selecting the dataset, the user would press the compute button and navigate to the Splines or Strengths and Weaknesses section to see the rendered plots as in Figure \ref{Splines&SW}.

Unlike the \texttt{airt} package, users can select an algorithm to highlight that algorithm's spline in the Splines section or remove the grey areas corresponding to the standard errors around each smoothing spline. In the Strengths and Weaknesses section, users can move a vertical slider on the left of the plot to change the epsilon value, which re-renders the plot.

\subsubsection{AIRT Shiny App Internals}
Internally, after the user uploads and presses compute, the dataset is passed to the different \texttt{airt} functions and cached. In this case, the server loads \texttt{classification\_cts} from their `Classification' example file selection, following the same \texttt{airt} workflow listed above.

However, to allow for more fine-grained controls of the plot generation, AIRT Shiny App typically doesn't use \texttt{autoplot} when UI controls are added. Because of the number of possible plots with UI controls, only the default plots are cached; for example, a Strengths and Weaknesses plot with \texttt{epsilon} set to 0. For other plots, there is a general flow of fetching the cached \texttt{cirtmodel()} and \texttt{latent\_trait\_analysis()}, and then using:

\begin{verbatim}
//in server.R
renderPlot({
    generate_plot(plottype, epsilon_value)
})
\end{verbatim}

\noindent to send the plot to the UI. This \texttt{generate\_plot} function is a wrapper around \texttt{autoplot} to allow code reuse.

In situations with UI controls, such as the Splines section, custom plotting functions are used instead. In the case of the Splines plot, \texttt{generate\_splines()} generates the standard Splines plot from cached \texttt{airt} functions, but when an algorithm is selected, \texttt{generate\_spline\_plot()} is used instead. This function takes the algorithm chosen from the UI and uses \texttt{gghighlight} to highlight that algorithm.

\section{Impact}

We have outlined the AIRT-Module, a tool that provides unique and accessible insight into evaluating the performance of an algorithm portfolio using Item Response Theory. This module assists users in making empirical-based decisions with easily digestible data visualisations, a streamlined workflow flow, and a choice between using an R Package or a Shiny App. 

For a given task such as image classification, as the space of test problems expands, different algorithms are typically proposed to tackle different types of instances. Thus, discovering complementary algorithms is important as they can be part of an algorithm portfolio capable of tackling diverse instances. The AIRT-Module assists in showcasing algorithm diversity by computing IRT-based algorithm metrics and visualizing their strengths and weaknesses. Furthermore, it aids reproducibility, an important aspect in AI research.

As of 21st of August 2024, the \texttt{airt} has over 23900 downloads on CRAN. Moreover, a tutorial on the AIRT Shiny App was conducted at The Genetic and Evolutionary Computation Conference 2024.

\section{Conclusions}
We have presented AIRT-Module, a two-component R ecosystem comprising an R package and a Shiny app for algorithm portfolio evaluation. AIRT-Module brings insights from IRT -- a suite of methods from educational psychometrics -- to algorithm evaluation. 
Our framework enables a detailed and comprehensive analysis of algorithm performance across diverse problem settings, contributing to a more nuanced understanding of their strengths and weaknesses. This tool enhances the capability to position algorithms within a state-of-the-art portfolio and identify their strengths and weaknesses, ultimately advancing AI research. Future work can explore expanding our framework to incorporate a higher dimensional latent trait and adapting it to handle new data types and evaluation metrics.

\bibliographystyle{unsrt}  
\bibliography{myBibliography}

\end{document}